# Exploratory evaluation of politeness in human-robot interaction

Shikhar Kumar, Eliran Itzhak, Samuel Olatunji, Vardit Sarne-Fleischmann, Noam Tractinsky, Galit Nimrod, Yael Edan

Ben-Gurion University of the Negev, Beer-Sheva, Israel

**Abstract.** Aiming to explore the impact of politeness on Human-Robot Interaction, this study tested varying levels of politeness in a human-robot collaborative table setting task. Polite behavior was designed based on Lakoff's politeness rules. A graphical user interface was developed for the interaction with the robot offering three levels of politeness, and an experiment was conducted with 20 older adults and 30 engineering students. Results indicated that the quality of interaction was influenced by politeness as participants significantly preferred the robot's polite mode. However, the older adults were less able to distinguish between the three politeness levels. Future studies should thus include pre-experiment training to increase older adults' familiarity with robotic technology. These studies should also include other permutations of Lakoff's politeness rules.

## 1      Introduction

Politeness has been an important ingredient of social interactions, helping in building coordination and avoiding conflict and confrontation between individuals. Politeness in the field of Human-Robot Interaction (HRI) has been scarcely explored for assistive robots that could be used in everyday life of people. Saleem et al. [1] developed politeness in a receptionist (humanoid robot) based on Brown and Levinson's theory [2] of politeness. Another research used animation of robotic interaction between human and robot (gatekeeper) to study the evaluation of politeness among diverse participants [3]. In the field of compliance with a medical robot, a humanoid robot was used in healthcare service setting [4]. These studies focused on a sporadic interaction between humans and service robots.

Unlike previous research, the focus of this work is to explore the effect of politeness in socially assistive robots designed for a home setting. In this work, a human-robot collaborative task was developed based on Lakoff's theory of politeness [5], which covers both linguistic and behavioral aspects of polite communication. Lakoff defined politeness as "a system of interpersonal relations designed to facilitate interaction by minimizing the potential for conflict and confrontation inherent in all human interchange" [5]. Lakoff's rule of politeness goes beyond Grice's seminal coop-erative principle [6]. It has three sub-rules: *Don't Impose*, i.e., do not intrude into other people's business; *Give Options*, i.e., give the other person options either to refuse or to accept our requests and desires; and *Be Friendly* – provide informal expressions to communicate solidarity feelings. In this work we focus on these sub-rules following [7], who found that polite interactive human computer interface (HCI) systems led to increased satisfaction, trust, and enjoyment.

## 2     Methods

The study included a robotic arm setting up the table with meat meal utensils or dairy meal utensils to a pre-set position in collaboration with a user.

### 2.1     System design

A kuka iiwa lbr 14 R820 7 degrees of freedom robot was employed in a table setting task (see Fig. 1). The robotic arm was programmed in Python and executed in ROS. A Graphical User Interface (GUI) was developed using Java. Since this was an exploratory study only three levels of politeness were designed: No politeness, 1 rule activated, and 3 rules employed. To distinguish between the three levels, green, blue and red buttons, respectively, were used to start each phase of the experiment corresponding to a different level selected randomly by the user.

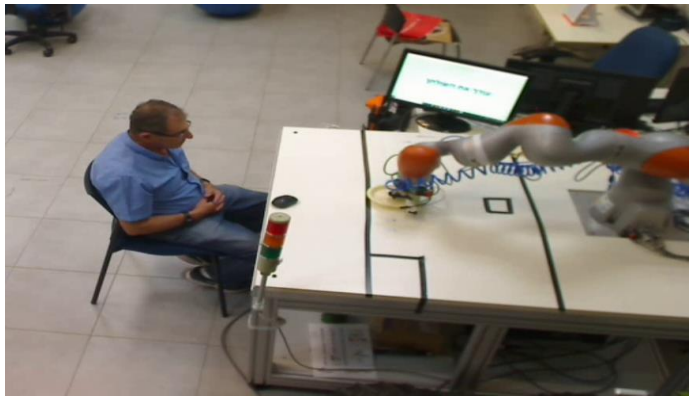

**Fig. 1.** Table setting robot

### 2.2     Politeness implementation

In the no-politeness condition, the robotic arm brought the dairy meal product at a pre-set position (in front of the user) with basic HRI. It displayed "setting the table" when robot was bringing utensils and also displayed "finish" when the task was completed. In the highest level of polite behavior (3 rules employed), the system welcomed and introduced itself to the user (i.e., it was *friendly*). The system did not im-pose the type of utensils. Rather, it asked the users when to set up the table (i.e., it *gave options*). In our study, the options were setting up the table now or waiting either one or two minutes. If the user selected the option of waiting, a timer with the selected time appeared on the screen and ran a down counter until zero. Then, the same window popped up and again gave the same 3 options. Only after selecting the "now" option, a new window inquired (i.e., *gave options*) about the user's preference regarding type of utensils. Here it gave the option of either "meat", "dairy" or "indifferent". The robot then set up the table with the selected option at pre-set positions (the indifference option was set to dairy utensils). Upon completion, the system asked the users for feedback, and asked if they were satisfied with the present table setup or wanted to change the utensils (if this option was selected, the robot would change the plates). After that, the system thanked the users for selecting its service and greeted them.

In the intermediate level of politeness only the *don't impose* rule was activated. This level was disregarded from the analysis due to reasons explained below.

### 2.3  Participants

The study was conducted with two different groups, belonging to the two ends of familiarity and openness towards technology: engineering students aged 24-29 (average=25.87, standard deviation= 5.621) and older adults aged 66-89 (average=73.85, standard deviation =5.235). The student group included 19 women and 11 men. The older adults group included 12 women and 8 men. The participants were recruited by advertising online and snowball sampling. Ethical procedures were approved by the Institutional Review Board.

### 2.4  Procedures and Measurement

The experiment began with providing the participants with brief explanation and instructions. After signing the consent form, they replied to a background questionnaire probing their age, gender, previous exposure to robotic technology, uses of computer, mobile, etc. They then moved to the experimental setup where they were asked to randomly choose between the colored buttons representing the politeness levels and started the first experiment trial. Following this, they chose another level (color) and executed the experiment until all three levels were completed. After each trial they completed a post-trial questionnaire in which answer options ranged from 1 ("completely disagree") to 5 ("completely agree"). The questionnaire's items pertained to the users' satisfaction (3 questions), trust (2 questions), and enjoyment (3 questions) of the interaction. The mean of the ratings given by the user was calculated and used as the subjective measures. Additionally, participants' heart rates were measured pre and post each trial, and the change in heart rate was calculated as the objective measure. This measure was used to measure the anxiety or excitement level of participants during the different level of politeness and during each trial based on previous HRI research which used this measure [8].

### 2.5 Analysis

Paired sample t-tests were conducted for each of the four dependent variables namely subjective measures and objective measure for a 0.0125 level of significance (corresponding to Bonferroni correction for 4 tests with p=.05). The independent variable was defined as politeness with two levels, i.e., polite behavior (3 rules employed) and no polite behavior. We disregarded the intermediate politeness level from the analysis since the older adults did not differentiate between the levels as detailed below. The null hypothesis is that politeness and no politeness are similar for all measures (subjective and objective).

## 3   Results and discussion

The mean of change in heart rate for first, second and third trial were -0.40, -0.67, and -0.03 for students and -14.75, -0.2 and -2.05 for older adults respectively. Older adults were probably anxious (or excited) to interact with the robot at the start of the experiment as observed in the sharp decrease in heart rate in the first encounter. With regard to politeness level, the change in heart rate was not significant for both the groups. The null hypothesis for this measure in both young and old participants was accepted.

The mean and standard deviation of scores of the participants' responses (Table 1) depicts that the polite behavior was preferred in both age groups. In all subjective measures for students the 'polite' level had significant difference over the 'no polite' level (satisfaction: $t(29) = -11.91$, $p = 1.08\text{e-}12$; trust: $t(29) = -4.12$, $p=2.91\text{e-}04$; enjoyment: $t(29) = -5.74$, $p = 3.27\text{e-}06$). However, for older adults' significant difference between them was not found (satisfaction: $t(19) = -1.93$, $p = 0.06$; trust: $t(19) = -1.08$, $p = 0.29$; enjoyment: $t(19) = -1.10$, $p = 0.28$).

In the post experiment questionnaire, 96.67% of the students but only 50.00% of the older adults noted they felt a difference between the scenarios. Yet, 76.67% of the students and 55.00% of the older adults preferred the polite behavior.

These results suggest that polite behavior is preferable in both the groups. It is important to note that in the 'no polite' level case, it was a one-way communication which may have an influence on the results in favor of polite behavior (i.e., there was less interaction). However, in the case of older adults these results call for further research. It is clear that the older adults were unfamiliar with the technology and this might have been the reason some of them had difficulties differentiating between the experimental conditions (the politeness levels). They may have also been too concen-trated on the robot itself, its motion, and the successful completion of the task to notice the differences between the two politeness levels.

**Table 1.** Mean score level of subjective measures

| Measures | Satisfaction | | Trust | | Enjoyment | |
|---|---|---|---|---|---|---|
| Behavior | No Polite | Polite | No Polite | Polite | No polite | Polite |
| Older adults | 2.82±1.05 | 3.25±0.89 | 3.80±0.99 | 4.05±0.82 | 3.73±1.18 | 3.98±0.80 |
| Students | 2.22±0.68 | 4.20±0.58 | 3.58±0.90 | 4.33±0.59 | 3.04±0.89 | 3.97±0.73 |

## 4  Conclusions

This exploratory study indicated that polite robot behavior is preferred among both older adults and students. However, the fact that only 50% of the older adults were able to differentiate between the levels of politeness suggests that they were overwhelmed with the technology during their first encounter. This was reflected by both the subjective and the objective measures. Future studies should therefore consider a training session with the robot to get the user familiar with the system before evaluating politeness. Yet, social robot designers should also consider designs that alleviate deleterious effects of first encounters with social robots in real-world settings where training is not feasible. Future work should also investigate the effects of intermediate levels of politeness based on Lakoff's theory and designing the 'no polite' condition more interactive to provide similar interaction conditions since interaction itself might influence the results. This would help in analyzing the pattern in which change of politeness level would affect the quality of interaction. This research should also be extended to a larger population. Additionally, the cultural context and user profile and background may have an effect on politeness preferences and expectations, and hence should be studied.

## Acknowledgments

This research was supported by the EU funded Innovative Training Network (ITN) in the Marie Skłodowska-Curie People Programme (Horizon2020): SOCRATES (Social Cognitive Robotics in a European Society training research network), grant agreement number 721619. Partial support was provided by Ben-Gurion University of the Negev Agricultural, Biological and Cognitive Robotics Initiative, and the Rabbi W. Gunther Plaut Chair in Manufacturing Engineering.